\pdfoutput=1

\documentclass[11pt]{article}

\usepackage[review]{ACL2023}

\usepackage{times}
\usepackage{latexsym}
\usepackage{booktabs}
\usepackage{graphicx}
\usepackage{listings}
\usepackage[T1]{fontenc}

\usepackage[utf8]{inputenc}

\usepackage{microtype}
\usepackage{todonotes}

\usepackage{inconsolata}

\title{Biomedical Entity Linking with Triple-aware Pre-Training}

\author{Xi Yan \\
 HITEC e.V.  \\
   Universität Hamburg \\
  \texttt{xi.yan@uni-hamburg.de} \\
  \And
  Cedric Möller  \\
  Universität Hamburg  \\
  \texttt{cedric.moeller@uni-hamburg.de} \\
  \AND
         Ricardo Usbeck \\ Universität Hamburg \\ricardo.usbeck@googlemail.com \\ }

\begin{document}
{\makeatletter\acl@finalcopytrue
  \maketitle
}
\begin{abstract}
Linking biomedical entities is an essential aspect in biomedical natural language processing tasks, such as text mining and question answering. However, a difficulty of linking the biomedical entities using current large language models (LLM) trained on a general corpus is that biomedical entities are scarcely distributed in texts and therefore have been rarely seen during training by the LLM. At the same time, those LLMs are not aware of high level semantic connection between different biomedical entities, which are useful in identifying similar concepts in different textual contexts. To cope with aforementioned problems, some recent works \cite{Yuan2022, neighBERT} focused on injecting knowledge graph information into LLMs. However, former methods either ignore the relational knowledge of the entities or lead to catastrophic forgetting. Therefore, we propose a novel framework to pre-train the powerful generative LLM by a corpus synthesized from a KG. In the evaluations we are unable to confirm the benefit of including synonym, description or relational information.

\end{abstract}

\section{Introduction}

Biomedical entity linking (EL) is a critical process in biomedical text mining that seeks to identify and associate relevant biological and medical entities mentioned in unstructured text with their corresponding identifiers in knowledge bases. Accurate recognition and linking of these entities are pivotal in promoting biomedical research, drug discovery, and personalized medicine \cite{primekg}. Although substantial progress has been made in recent years, there is an ongoing need for refining methods and techniques employed for entity linking in the biomedical domain.

In this report, we present a novel approach that integrates linearized triples into the biomedical entity linking process while reevaluating the inclusion of synonym information. Our proposed method linearizes triples and considers them during the pre-training step. In past studies, synonym information, which involves using alternative names or terminologies for the same biomedical entity, has been suggested as a means to enhance entity linking when used during pre-training  \cite{Yuan2022, cross_entity_attention}. Our study aims to build upon this existing knowledge by integrating both strategies and assessing their impact on performance.

Despite the reported benefits of synonym information in prior studies, our analysis of this approach, combined with the introduction of linearized triples \cite{linearilization}, yielded different results. We find that incorporating linearized triples only lead to minimal improvements in our entity linking model's performance. Moreover, we are unable to confirm the purported advantages of including synonym information in our experiments, which stands in contrast to the findings of previous literature.

We highlight the limitations of our study and suggest possible avenues for future research to further advance biomedical entity linking techniques by building on our work with linearized triples and reevaluating synonym information.

\section{Related work}
Entity Linking has a long history of research. Recent methods can be categorized into two types. First, discriminative methods that are based on the bi-encoder / cross-encoder pairing~\cite{Wu2020, Logeswaran2019, Ayoola2022}. Both encoders are commonly BERT-like models. The bi-encoder encodes the description of each entity and matches it to the text by using an approximate nearest neighbor search. This is important as the next step, the cross-encoding, is expensive. Here, those neighbors are reranked by applying a cross-encoder to the concatenation of both, the input text and the entity description. The highest-ranked entity is then the final linked one. 
In the biomedical domain, the works by ~\citet{Angell2021}, ~\citet{Varma2021},~\citet{Agarwal2021} and \citet{Bhowmik2021} fall into this category.

Another type of entity linker is based on generative models~\cite{Cao2021, Cao2022, cross_entity_attention}. Here, instead of using some external description of an entity, the whole model memorizes the KG during training. The linked entity is then directly generated by the model. Such methods skip the problem of mining negatives which are crucial for a good performance of bi-encoder-based methods. 
Only the work by~\citet{Yuan2022} is based on such methods in the biomedical domain. 
As generative models lack the ability to incorporate external information, they alleviate this problem by
introducing a pre-training stage where syntactical information from a knowledge graph is learned. This is especially important in the biomedical domain as entities often own a large variety of synonyms.
We build upon their work by extending the pre-training regime to the inclusion of triple information.

\section{Method}
\subsection{Task definition}
Given are a text $t$, a set of marked mentions $M_t$ in the text and a KG $\mathcal{G}=(\mathcal{E}, \mathcal{R}, E)$. The KG consists of a set of entities $\mathcal{E}$, a set of relations $\mathcal{R}$ and a set of edges composed of head entity, relation and tail entity $E \subseteq (\mathcal{E} \times \mathcal{R} \times \mathcal{E})$. 
The task is to identify the subset of entities $E_t \subseteq \mathcal{E}$ which the mentions  $M_t$ are referring to. 

\subsection{Model}
In the vein of the work by  
~\citet{Cao2021}, we model the problem as a sequence-to-sequence generation task. The input to the generative model is text and the output are the generated entity identifiers in the corresponding KGs. Similar to other works~\cite{Cao2021,Cao2022, Yuan2022}, we consider the definition of the concepts in the corresponding KGs as the unique textual representation of each concept.  

\subsection{Pre-training}
We linearize the information from synonym and triples in the pre-training stage. They are linearized into a synthesized corpora before feeding into the BART. We have tested 2 different settings for converting the triples, namely \textbf{line-by-line} and \textbf{all-in-one}.

In terms of the \textbf{synonym information}, we follow the setting by~\citet{Yuan2022}.
We first extract the description of the entity and convert it to a text of the following form:
\begin{lstlisting}[basicstyle=\small]
    [BOS][ST]$s^a_e$[ET] is defined as $c_e$[EOS]
\end{lstlisting}
\label{example_1}
Here, $s^a_e$ stands for the synonym $a$ and $c_e$ for the description of entity $e$. 
This would be the input to the encoder of the generative model. 

As an output the model has to generate: 
\begin{lstlisting}[basicstyle=\normalsize]
    [BOS] $s_e^a$ is $s_e^b$ [EOS]
\end{lstlisting}
This lets the model learn the connection between the different synonyms of the same entity.

Based on that, we introduce an additional pre-training step to incorporate more semantic information by utilising \textbf{triple information} from the underlying knowledge graph. A triple is of the form $<e, r, e'>$ which describes that a relationship $r$ holds between entity $e$ and $e'$.
The input is here the same as for the synonym information. The output is of the form:
\begin{lstlisting}[basicstyle=\normalsize]
    [BOS] $s_e^a$ $l_r$ $s^b_{e'}$ [EOS]
\end{lstlisting}
$l_r$ is here the label of relation $r$. We denote this \textbf{line-by-line}

\begin{figure*}[!htb]
    \centering
    \includegraphics[width=0.95\textwidth]{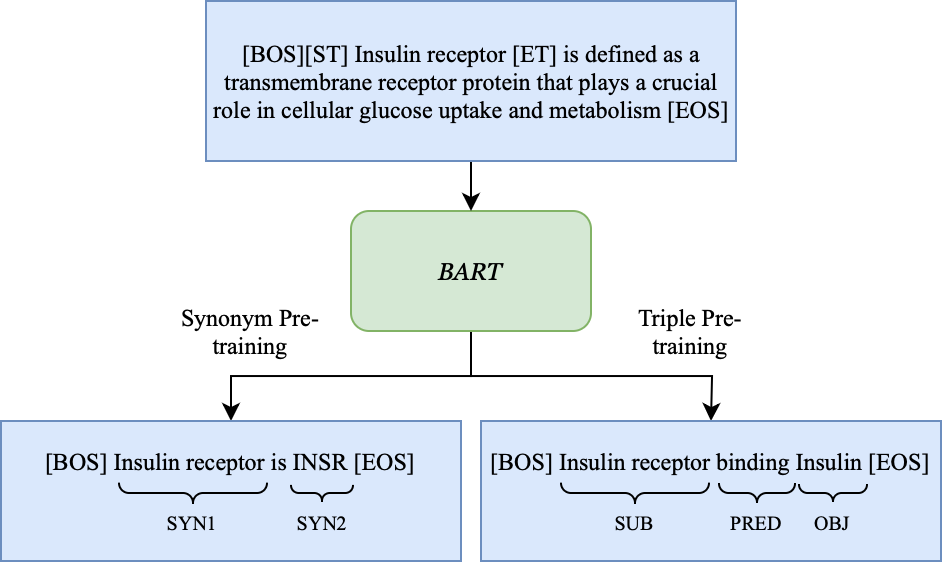}
    \caption{An overall workflow of our framework. We adopt different textualization formats for synonym information and triples. Both are included in the pre-training stage.}
    \label{fig:pretraining}
\end{figure*}

Furthermore, we experimented with an \textbf{all-in-one} pre-training approach of the form:
\begin{lstlisting}[basicstyle=\normalsize]
    [BOS] $s^a_e$ $l_{r_1}$ $s^{b_1}_{e_1}$ $\dots$ $l_{r_n}$ $s^{b_n}_{e_n}$ [EOS]
\end{lstlisting}

See Figure~\ref{fig:pretraining} for an overview of the pre-training.

\subsection{Fine-tuning}
During fine-tuning, the model is trained for the actual entity linking task.
The input to the generative model is the unlabelled biomedical text.
To generate the linked entities, each mention is included in a template  as follows:
\begin{lstlisting}[basicstyle=\normalsize]
    [BOS] $m_i$ is $s^a_e$ [EOS]
\end{lstlisting}
The model then generates the entity identifier after the token "\texttt{is}".
Similar to the work by \citet{Yuan2022}, we choose the synonym which is syntactically close to the corresponding mention in the text as the target entity identifier during fine-tuning.

The generated entity identifier is mapped back to the concrete entity in the final step via a lookup table. During inference, we restrict the possible output space by limiting it to the available entity names and synonyms.

\begin{figure}[!htb]
    \centering
    \includegraphics[width=0.95\linewidth]{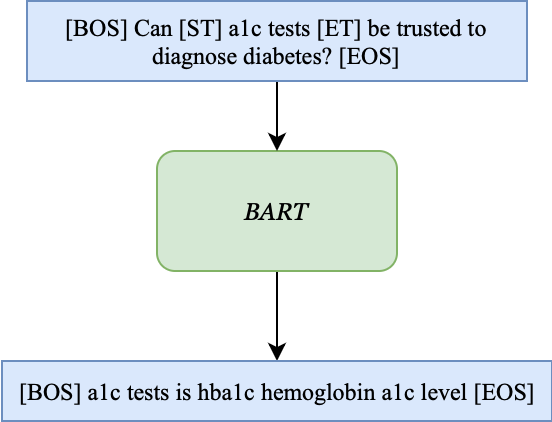}
    \caption{ An overview of the fine-tuning stage}
    \label{fig:finetuning}
\end{figure}

See Figure~\ref{fig:finetuning} for an overview of the pre-training.

\section{Evaluation}
\subsection{Pre-training Strategy} 

We use a synthesized corpus composed of triples, synonyms and descriptions from UMLS. More specifically, we decide to use a subset of UMLS, i.e., st21pv \cite{subset_UMLS}. It is a well-connected KG with information about concept definitions and synonyms. Specifically, 160K out of 2.37M concepts have definitions, 1.11M concepts have several synonyms and 68K concepts are connected to on average 8 triples. During the pre-training step, we construct samples by iterating through each concept's synonyms and triples. Each concept is densely connected and the distribution of the number of triples a concept is connected to is skewed. For instance, some "popular" concepts are connected to over 1000 triples, while some are connected to only 1 triple. To avoid the class imbalance, we sample the included triples based on the relation frequencies.

To train the model with KG information, we linearize triples.
Linearization refers to a special type of technique on converting graph to text, i.e., converting triples to one/more sentences which serve as input of the LLM. 




We sample the included triples based on the relation frequencies. First, we gather the occurrence frequency of all relations in the KB by counting the number of triples this relation is connected to. We define the probability ($P_{r}$) of a relation $r$ to be negatively related to the frequency. Then, for each concept in the KG, we collect its connected triples and segment the triples into different groups based on their relation $r$. 

Both settings are trained under the same experiment setting  with a batch size of 128. We save the best model within 12 training epochs. 

\subsubsection{Fine-tuning}
The model is fine-tuned on two established datasets, namely BC5CDR (\cite{bc5cdr}) and NCBI (\cite{ncbi}). Those entity linking datasets are constructed on subsets of UMLS, making them perfect choices to test our model's performance on. Among the datasets, NCBI and BC5CDR are generated by annotating PubMed papers. On the other hand, NCBI and BC5CDR are annotated against  Medical Subject Headings (MeSH) - a terminology knowledge graph for indexing and cataloging of biomedical information.

The statistics of the four datasets are exhibited in Table~\ref{table: dataset stat} below. As we can see, NCBI and BC5CDR (annotated on academic text) are smaller in size. Also NCBI and BC5CDR are dense in terms of the target entities they contain (14,967 and 268,162).

\begin{table*}[!htb]
\centering
\begin{tabular}{l c c c c}
\toprule
Nums & NCBI  & BC5CDR \\
\midrule
  Train   & 5,784    & 9,285                  \\
   Dev  & 787  & 9,515        \\
 Test & 960   & 9,654               \\
 Entities & 14,967 & 268,162 \\
\bottomrule
\end{tabular}
\caption{Numbers of the samples in the training, development and test set}
\label{table: dataset stat}
\end{table*}

BART-large ~\cite{BART} is chosen as the generative model as it has been an established benchmark model for such tasks.

\subsection{Results}
We assess the performance of four distinct models in the entity linking task, including two of our own models, each pre-trained via either a line-by-line or all-in-one strategy, a synonym pre-trained model from \cite{Yuan2022} (denoted Syn-Only), and a basic BART model. We also include the recent papers which pretrains BART on biomedical domain \cite{biobart} before finetuned on biomedical entity linking datasets and ResCNN \cite{lai-etal-2021-bert-might} which achieves state-of-the-art results on various biomedical EL datasets. Each model undergoes fine-tuning specific to the entity linking task. Recall@1 for each model are presented in Tables \ref{table:result1} and \ref{table:result2}. We limit ourselves to Recall@1 to follow the common practice when measuring entity linking performance without named entity recognition.  The best-performing metrics are emphasized in bold.




\begin{table}[!htb]
\centering
\begin{tabular}{lllll}
\toprule
                                                 & BC5CDR                                  & NCBI                   \\ \midrule

Syn-Only                                          & 93.3\%                                 & 91.9\%                 \\ \midrule
Syn-Only                                             & 92.68\%                              & 89.45\%                \\ 
All-in-one                                           & 92.86\%                               & 88.43\%                \\ 

Line-by-line                                         & 92.66\%                           & 90.00\%                \\ 
BART        & 92.58\%                               & 89.06\%   \\ 
BioBART-Large       &  93.01\%      & 89.27\%                           \\ 
BioBART-Base        & \textbf{93.26}\%      & 89.40\%                           \\ 
ResCNN        & 91.7 \%      & \textbf{92.4}\%                           \\ 

\bottomrule
\end{tabular}
\caption{Recall@1 on BC5CDR and NCBI, which are PubMed articles annotated against MESH.}
\label{table:result2}
\end{table}

\subsection{Analysis}
Based on the table, our triple injection framework exceeds the BART baseline on the 2 Wbenchmarks datasets. On BC5CDR and NCBI, the gain compared to BART  is around 0.2\% and 0.5\%.

\textbf{Does triple injection enhance model's capacity to link to the correct entity?} The answer is yes, since over 4 datasets, the All-in-one or Line-by-line variants outperform the variant that was not trained on the linearized corpora for around 1\% (Recall@1).

\section{Conclusion}
Our study seek to improve biomedical entity linking through the integration of linearized triples and synonym information. However, contrary to expectation, the incorporation of these elements leads to only minimal improvements in our EL model performance.

 In conclusion, our study underscores the complexities of biomedical EL and prompts the need for more sophisticated approaches to improve its accuracy. A possible future extension of this work could be to explore more sophisticated methods to instruct the LLMs to learn external knowledge, such that the knowledge is injected in an efficient way which benefits the models in downstream tasks. For instance, by incorporating the KG information not just in a linearized manner but by exploiting the graph-structure with Graph Neural Networks \cite{GNNs}, mutliple methods could be further developed.   
 
This work has been partially supported by grants for the DFG project NFDI4DataScience project (DFG project no. 460234259) and by the Federal Ministry for Economics and Climate Action in the project CoyPu (project number 01MK21007G).
\bibliography{acl2023.bbl}
\bibliographystyle{acl_natbib}

\end{document}